\documentclass[letterpaper, 10 pt, conference]{ieeeconf}  

\IEEEoverridecommandlockouts                              

\usepackage{caption}
\usepackage{blindtext}
\usepackage{graphicx}
\usepackage{hyperref}
\usepackage{amsmath}
\usepackage{amssymb}
\usepackage{tabularx}
\usepackage{subcaption}
\usepackage[acronym]{glossaries}
\usepackage{wrapfig,lipsum,booktabs}
\usepackage{siunitx}
\usepackage{float}
\usepackage{multicol}

\glsdisablehyper
\newacronym{mse}{MSE}{Mean Squared Error}
\newacronym{mlp}{MLP}{Multi-Layer Perceptron}
\newacronym{svd}{SVD}{Singular Value Decomposition}
\newacronym{icp}{ICP}{Iterative Closest Point}
\newacronym{slam}{SLAM}{Simultaneous Localisation and Mapping}
\newacronym{ransac}{RANSAC}{Random Sample Consensus}
\newacronym{dl}{DL}{Deep Learning}
\newacronym{gat}{GAT}{Graph Attention Network}
\newacronym{imu}{IMU}{Inertial Measurement Unit}
\newacronym{gnn}{GNN}{Graph Neural Network}
\newacronym{rte}{RTE}{Relative Translational Error}
\newacronym{rre}{RRE}{Relative Rotational Error}
\newacronym{gcn}{GCN}{Graph Convolution Network}

\usepackage{xcolor}

\usepackage{siunitx}
\usepackage{mathtools}
\usepackage{physics}
\DeclarePairedDelimiterX\set[1]\lbrace\rbrace{\def\given{\;\delimsize\vert\;}#1}

\usepackage{multirow}

\usepackage{cuted}

\usepackage{cleveref}
\crefname{table}{Tab.}{Tabs.}
\crefname{figure}{Fig.}{Figs.}
\crefname{section}{Sec.}{Secs.}
\crefname{equation}{Eq.}{Eqs.}
\usepackage[hang,flushmargin]{footmisc}

\begin{document}

\title{\huge \bf
SEM-GAT: Explainable Semantic Pose Estimation using Learned Graph Attention
}

\author{Efimia Panagiotaki$^1$, Daniele De Martini$^2$, Georgi Pramatarov$^2$, Matthew Gadd$^2$, Lars Kunze$^1$\\
$^1$Cognitive Robotics Group and $^2$Mobile Robotics Group, Oxford Robotics Institute, \\ Department of Engineering Science, University of Oxford, UK\\
\texttt{\{efimia,daniele,lars\}@robots.ox.ac.uk}
\thanks{Project code: github.com/cognitive-robots/SEM-GAT\vspace{0.10cm}}
\thanks{This project was supported by a Google DeepMind Engineering Science Scholarship, EPSRC projects RAILS (EP/W011344/1), and Programme Grant ``From Sensing to Collaboration'' (EP/V000748/1).
}
}
\maketitle

\begin{figure*}
\vspace{1.5mm}
\centering
\includegraphics[width=\textwidth]{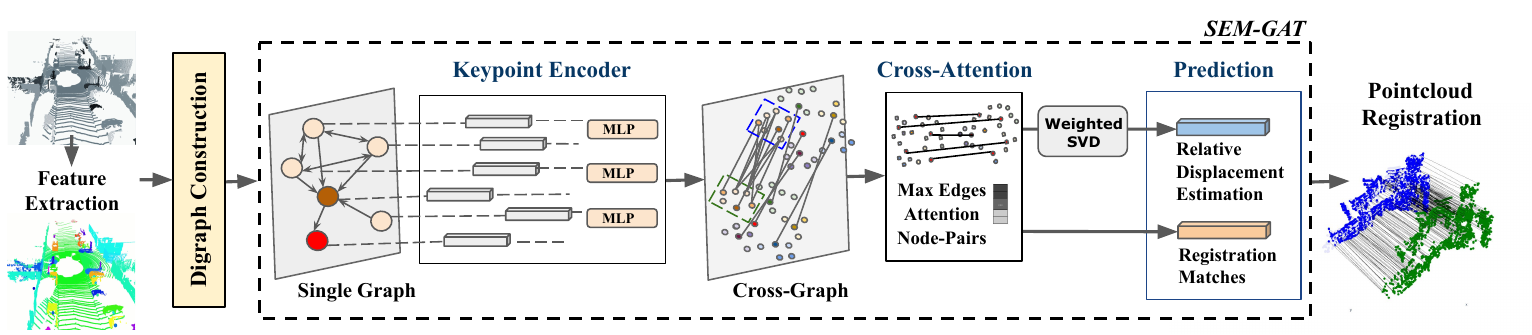}
\caption{
We extract a morphology- and semantics-informed graph structure (see \cref{sec:graph_construction}) and generate embedding representations through local-neighbourhood aggregation (see \cref{sec:node embeddings}).
We then use cross-attention (\cref{sec:cross_graph_attention}) to assign confidence scores to point-to-point matching candidates for pointcloud registration and employ a weighted SVD module to estimate the relative pose between the pointclouds, represented as nodes in the cross-graph structure (\cref{sec:displacement_estimation}).
}
\vspace{-0.5cm}
\label{fig:architecture}
\end{figure*}

\begin{abstract}
This paper proposes a \gls{gnn}-based method for exploiting semantics and local geometry to guide the identification of reliable pointcloud registration candidates.
Semantic and morphological features of the environment serve as key reference points for registration, enabling accurate lidar-based pose estimation.
Our novel lightweight static graph structure informs our attention-based node aggregation network by identifying semantic-instance relationships, acting as an inductive bias to significantly reduce the computational burden of pointcloud registration. 
By connecting candidate nodes and exploiting cross-graph attention, we identify confidence scores for all potential registration correspondences and estimate the displacement between pointcloud scans.
Our pipeline enables introspective analysis of the model's performance by correlating it with the individual contributions of local structures in the environment, providing valuable insights into the system's behaviour. We test our method on the KITTI odometry dataset, achieving competitive accuracy compared to benchmark methods and a higher track smoothness while relying on significantly fewer network parameters.
\end{abstract}
\begin{keywords}
Pointcloud registration, pose estimation, graph neural networks, attention, eXplainable AI (XAI)
\end{keywords}
\glsresetall
\section{Introduction}\label{sec:introduction}
\noindent Robust and reliable localisation is a key component in the software pipeline of autonomous mobile systems, ensuring their safe navigation and task execution. Localisation can be split into the sub-tasks of place recognition and pose estimation, both equally important to the autonomous pipeline. 

Lidar-based pose estimation relies on accurate pointcloud registration to estimate the displacement between two pointclouds and eventually align them. Registration is commonly achieved by extracting local or global descriptors, either handcrafted \cite{He2016M2DP,rusu2009fpfh} or learned \cite{sun2020dagc,Zhang2019pcan}, to identify distinctive keypoint matches. There has been an increase in semantic- and segments-based descriptor methodologies in recent years \cite{Dube2017SegMatch,Li2021b}, offering a good compromise between local and global descriptors combining the advantages of both.

Due to their structural flexibility and capacity to conceptualise high-level information, \glspl{gnn} are increasingly used in place recognition and odometry estimation tasks. However, state-of-the-art approaches either use fully-connected or dynamic graphs without explicitly exploiting important relational semantic information \cite{Wang2019DCP,fischer2021stickypillars}, or, if relying on semantics, focus solely on the instances' centroids without adequately exploiting local structural information \cite{Li2021b,Li2021SSC}.


In this work, we propose SEM-GAT, a learned graph-attention-based semantic pose estimation approach that exploits local structural \textit{and} contextual information through a novel semantic-morphological static graph structure, to guide the model's predictions. SEM-GAT uses a \gls{gnn} architecture relying on attention weights to identify pointcloud registration candidates and estimate the relative displacement between them. Our approach provides an explainable methodology that utilises high-level information to optimise registration accuracy while identifying key structures in the pointclouds. This pipeline enables us to analyse the complexity of the environment by investigating relationships between scene elements and their contribution to pose estimation, further explaining the model's behaviour.

Our key contributions are summarised as follows:
\begin{itemize}
    \item A novel static graph architecture guided by local geometry and semantics in pointclouds;
    \item A learned \gls{gnn} pose estimation pipeline based on attention for identifying point-wise registration candidates;
    \item An explainability analysis of the predictions examining the correlation between semantics and morphology in registration performance.
\end{itemize}
We validate our approach on the KITTI odometry dataset \cite{6248074}, achieving competitive results in pointcloud registration accuracy compared to benchmarks while facilitating greater model introspection and requiring fewer learned parameters.

\section{Related Work}\label{sec:related_work}

The classical approach for local pointcloud registration is \gls{icp} and its variants \cite{besl1992method,Rusinkiewicz2001,Segal2009,Zhang2022}, which iteratively align two pointclouds by relying on nearest neighbours to find point-to-point correspondences between closest points. Scan matching techniques use \gls{icp} as prior for \gls{slam} \cite{Mendes2016,Borrmann2008}. As these methods rely on point-wise distance to find registration correspondences, they do not exploit valuable pointcloud characteristics to extract descriptive keypoints and are less robust to noise.

Global registration approaches that rely on keypoint-matching, on the other hand, depend on consistent keypoint-extraction techniques~\cite{li2019usip} and distinctive hand-crafted~\cite{rusu2009fpfh} or learned~\cite{choy2019fcgf} feature descriptors. To extract registration candidates, they employ outlier rejection methods~\cite{rusu20113d, yang2021teaser}, most commonly \gls{ransac}~\cite{Fischler1981}. As these traditional methods mainly depend on spatial closeness and are not differentiable, the advent of \gls{dl} has inspired new registration techniques that can be trained in an end-to-end fashion.
DCP \cite{Wang2019c}, IDAM \cite{Li2020b}, and DeepICP~\cite{Lu2019} are differentiable pointcloud registration approaches that find similarities in feature space.
However, these techniques have only been tested on small datasets as they are either too computationally intensive or aggressive at rejecting keypoints. 

Other differentiable approaches such as DGR~\cite{choy2020dgr}, PCAM~\cite{cao2021pcam}, and GeoTransformer~\cite{qin2022geo} aim to predict confidence scores of keypoint matches, thus avoiding the use of iterative outlier rejection via \gls{ransac}. Similarly, in the graph domain, methods like SuperGlue \cite{Sarlin2020SuperGlue}, StickyPillars \cite{fischer2021stickypillars}, and MDGAT \cite{shi2021keypoint} rely on learned attention weights to extract matching correspondences. These methods use dynamic graphs extracted from local neighbourhoods to find correspondence candidates without considering morphological and semantic elements to inform registration and reduce the size of the graphs. SGPR~\cite{Kong2020sgpr} was the first method to exploit a semantic graph representation; however, SGPR uses only the centroids of the detected semantic instances to build the graphs, discarding important information about their geometry and structure. 

Conversely, our proposed methodology utilises \textit{both} semantic and geometric information from lidar scans to extract relationships between points, informing the graph generation and registration process. In our approach, we exploit morphological and high-level semantic characteristics of local pointcloud structures
to build a graph representation and identify registration candidates. We then employ an attention-based \gls{gnn} model to learn strong registration matches from keypoints generated after aggregating local neighbourhood nodes in the graphs.

\section{Overview}\label{sec:overview}

\Cref{fig:architecture} visualises the overall pipeline. We can distinguish two main parts: the digraph generation process and the \gls{gnn}-based registration pipeline, discussed in \cref{sec:graph_construction} and \cref{sec:registration} respectively.
Our graph generation approach is guided by geometric and semantic information in the pointclouds and our cross-graph registration pipeline is designed to exploit that structure to learn confidence scores for each registration candidate. In \cref{sec:results_introspect}, we further explain the performance of our model in correlation with the semantic features present.

\subsection{Problem definition and notation}
Let $P_k : \set{\mathbf{p}_i \given \mathbf{p}_i \in \mathbb{R}^3}$ be a pointcloud at time $k$, and $P_l : \set{\mathbf{p}_j \given \mathbf{p}_j \in \mathbb{R}^3}$ be a pointcloud at time $l$.

We assume $P_l$ is transformed from $P_k$ by a rigid transformation $[\mathbf{R}_{k,l}|\mathbf{t}_{k,l}]$ where $\mathbf{R}_{k,l} \in \mathbb{SO}(3)$ corresponds to rotation and $\mathbf{t}_{k,l} \in \mathbb{R}^3$ corresponds to translation. 
We aim to discover such transformation corresponding to the displacement between the two pointclouds.
In our specific case, we constrain ourselves to odometry, i.e. $l = k + 1$; yet, our formulation is generic, and as such we will use $k$ and $l$ throughout the discussion.

Given a set of matches between points, denoted as $M_{k,l} : \set{(\mathbf{p}_i, \mathbf{p_j}) \given \mathbf{p}_i \in P_k, \mathbf{p}_j \in P_l}$, we solve the pose estimation problem by applying a weighted \gls{svd} approach. The weights in \gls{svd} correspond to confidence scores for each $M_{k,l}$ when aligning the pointclouds. Thus, computing of the transformation $[\mathbf{R}_{k,l}|\mathbf{t}_{k,l}]$ is reduced to finding the matches $M_{k,l}$ with the maximum confidence scores.

We further aim to explain the model's behaviour by analysing the score assignment process in correlation with structures in the environment. In particular, we investigate using such confidence scores as importance indicators for local semantics and geometry.

\section{Graph Construction}\label{sec:graph_construction}

We propose a novel graph representation preserving key information and relationships within semantic instances to guide the discovery of registration matches. By exploiting local topology and semantics, we generate concise graph representations downsampling the pointclouds according to the semantic relationships identified between points.

\subsection{Feature extraction}\label{sec:feature_extraction}
For each point $\mathbf{p}_i \in P_k$ -- and similarly for each $\mathbf{p}_j \in P_l$ -- we extract information about its semantic and local geometry characterisation to construct the graphs.

\subsubsection{Semantics}
Let $\mathbb{S}$ be a set of semantic classes and $\mathbb{O}_k$ be the semantic instances of $P_k$, i.e. local regions of the scan with the same semantic class.
To each point $\mathbf{p}_i$, we assign a semantic label $s_i \in \mathbb{S}$.
We then extract the instances $\mathbb{O}_k$ by clustering the closest points in each semantic class $s_i$ according to spatial distance, constrained by sets of minimum cluster sizes and maximum clustering tolerances defined for each semantic category similar to \cite{Kong2020sgpr}.
The set of instances can then be defined as $\mathbb{O}_k: \set{ O_m \given O_m \subset P_k, O_m \cap O_n = \emptyset ~\forall m, n, s_o = s_q ~\forall \mathbf{p}_o, \mathbf{p}_q \in O_m}$.

Each instance $O_m$ is associated with a unique instance id $o_m$ and its centroid $C_m$. The centroid of each instance is extracted by computing the mean of the coordinates of all points in the cluster, $\mathbf{p}_i \in O_m$.
Semantic graph-based methods \cite{Kong2020sgpr,pramatarov2022boxgraph} discard the points and only keep the centroids to generate the graph. Instead, we use the morphology of the pointcloud \textit{and} the centroids to generate keypoints for registration considering the structural information of the instances.
%

\subsubsection{Local Geometry}
For each point $\mathbf{p}_i \in P_k$ -- and similarly $\mathbf{p}_j \in P_l$ -- we calculate the curvature $c_i$ of its local neighbourhood and classify it as either \texttt{corner} or \texttt{surface} point, similar to LOAM \cite{zhang2014loam}, as follows:
\begin{equation}
    c_i = {1 \over {|S|~||\mathbf{p}_{i}||}} ||\sum_{i^{\prime} \in S, i^{\prime} \neq i}(\mathbf{p}_{i}-\mathbf{p}_{i^{\prime}})||
\end{equation}
where $S$ is the set of consecutive points of $\mathbf{p}_i$.
LOAM uses this classification to downsample the pointcloud following a set of handcrafted rules categorising the points as inliers and outliers for registration.
We, instead, use the $c$ score to add context to the graph construction by assigning a geometric class of either \texttt{corner} or \texttt{surface} to each point.

\subsection{Single-Graph Construction}
\label{sec:singgraph}
\begin{figure}[t]
\centering
\includegraphics[width=0.9\columnwidth]{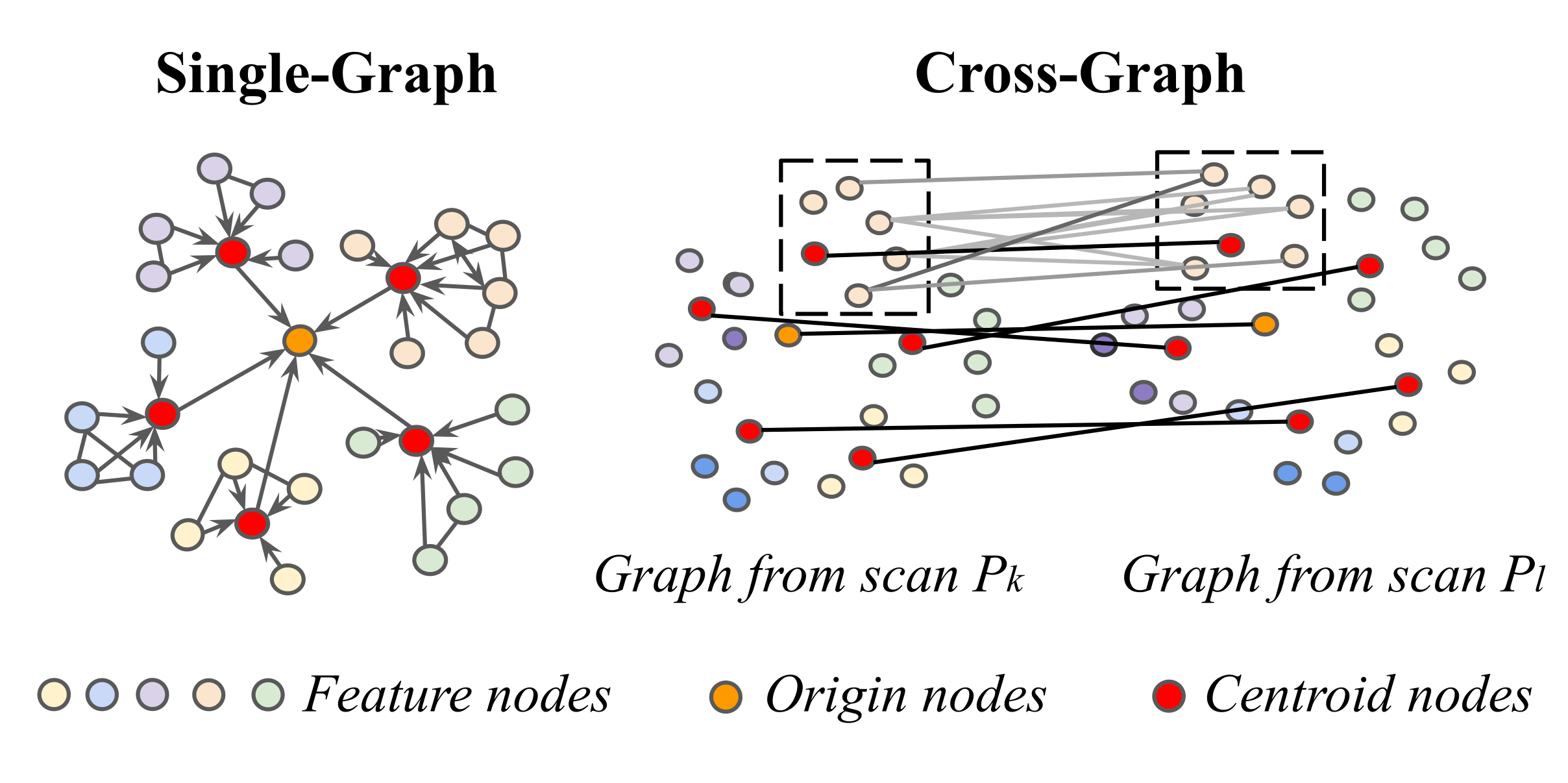}
\caption{Graph generation. (Left) Feature nodes in each instance are linked together and to their corresponding centroid, which is then connected to the origin. (Right) Same-class nodes in corresponding instances found in scan pairs are linked bidirectionally to form a cross-graph structure.
}
\label{fig:graphs}
\vspace{-3mm}
\end{figure}

Once the features for each point $\mathbf{p}_i \in P_k$ have been extracted, we use its geometric and semantic characterisation to construct the semantic graph $G_k$, as in \cref{fig:graphs}.

We define $G_k$ as $G_k = \langle N_k, E_k \rangle$, where $N_k$ and $E_k$ represent the set of nodes and edges respectively.
The graph $G_k$ is heterogeneous containing nodes of different types.
Each point can belong to one of the following classes: $F = \{$\texttt{origin}, \texttt{centroid}, \texttt{corner}, \texttt{surface}$\}$.
While the last three correspond to corner and surface points in the pointcloud and the instance centroids as defined above, the \texttt{origin} corresponds to a central node at the origin of the lidar scan.

Each node $n_i \in N_k$ is represented as a feature vector $n_i = [x_i, y_i, z_i, s_i, o_i, f_i]$,  describing its properties as:
\begin{itemize}
\item $x, y, z$ position: the 3D coordinates of the node in the lidar frame of reference;
\item Semantic id $s \in \mathbb{S}$: the semantic category of the node;
\item Instance id $o \in \mathbb{O}_k$: a unique identifier assigned based on the node's instance;
\item Feature id $f \in F$: a label indicating the class of the feature point.
\end{itemize}

To build the graph, we define a grammar corresponding to a set of rules representing semantic relationships as edges $E_k$ between nodes $N_k$. In each instance:
\begin{itemize}
\item Corner and surface nodes are connected to the centroid node, direction from corner/surface node to the centroid;
\item Each centroid node is connected to the origin, direction from the centroid to the origin;
\item Close points of the same category, corner or surface, are connected bidirectionally. 
\end{itemize}

This way, we form a hierarchy of connections to semantically guide the message passing in the \gls{gnn}.

\subsection{Cross-Graph Construction}\label{sec:graph_pair}

Given two pointcloud graphs $G_k$ and $G_l$ generated from the pointclouds $P_k$ and $P_l$, we construct a cross-graph structure $\Gamma_{k,l}$, defined as $\Gamma_{k,l} = \langle (N_k \cup N_l), (E_k \cup E_l \cup E_{k,l}) \rangle$, where $N_k$, $N_l$, $E_k$, and $E_l$ are the nodes and edges of the two graphs, and $E_{k,l}$ are edges that connect them.
\Cref{fig:graphs} shows an example visually. The edges $E_{k,l}$ between the nodes in the cross-graph connect registration candidates.

To populate $E_{k,l}$, given the relatively high frequency of the lidar in correlation with the vehicle's speed, we expect consecutive scans to have a relatively small displacement between them. Therefore, to identify corresponding instances, we find the closest centroids that belong in the same semantic class within a 3  \si{\meter} distance threshold. We then connect same-class feature points in corresponding instances within the same 3 \si{\meter} threshold. This method discards points that do not have a corresponding matching candidate due to their local geometric characteristics.

\subsection{Cross-Graph Pruning}\label{sec:graph_pruning}
\begin{figure}[h]
\vspace{2.2mm}
    \centering
    \includegraphics[width=\columnwidth]{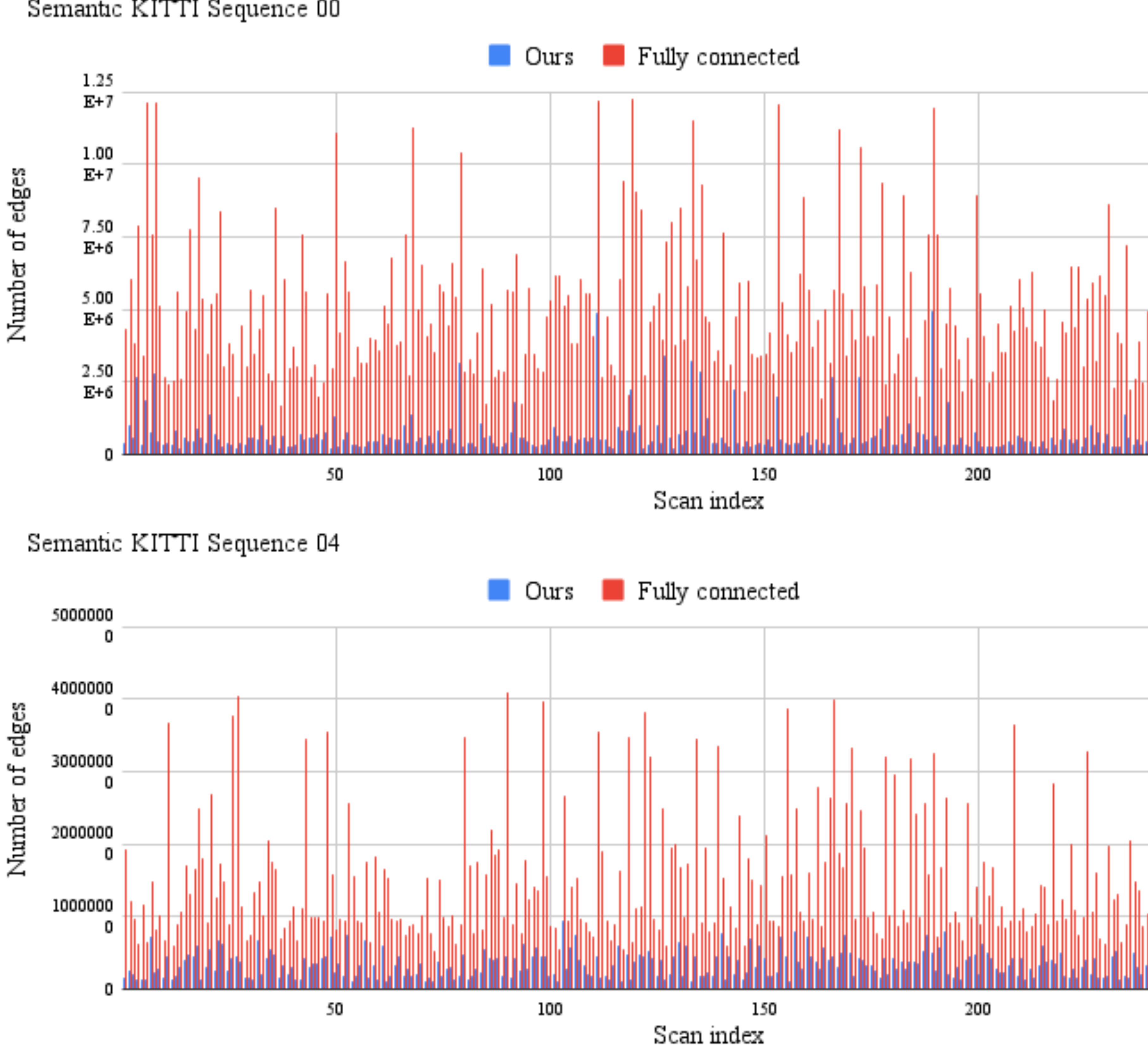}
    \caption{Number of edges per scan in sequences 00 and 04 of SemanticKITTI \cite{Behley2019}. Our graphs (blue) rely on significantly fewer edges for registration compared to a fully-connected graph with the same number of nodes (red).}    \label{fig:edges_number}
    \vspace{-3mm}
\end{figure}

As the two graphs are linked, feature nodes or instance centroids can be semantically unrelated.
We prune such nodes and edges from the cross-graph $\Gamma_{k,l}$ to speed up training and inference.
\Cref{fig:edges_number} depicts the resulting number of edges of $\Gamma_{k,l}$ compared to a fully-connected (FC) graph structure which is commonly used in attention-based \glspl{gnn} \cite{joshi2020transformers, murnanegraph}. In particular, our method utilises $89.49\%$ and $76.76\%$ less edges in sequences 00 and 04 of SemanticKITTI \cite{Behley2019}, respectively, compared to fully-connected approaches. With this substantial reduction in the number of edges, we achieve a lightweight graph structure, resulting in a significant decrease in memory requirements.

\section{SEM-GAT Network} \label{sec:registration}
Our proposed \gls{gnn} model learns to aggregate the nodes within the single-graphs, $G_k$ and $G_l$, and estimates confidence scores for each edge in $E_{k,l}$. These scores represent the likelihood of an edge connecting a point-to-point registration match and are then fed into a weighted \gls{svd} to estimate the offset between the two single-graph structures, approximating the matches set $M_{k,l}$.

The model can be split into three modules: 1) a keypoint encoder for single-graphs, 2) cross-graph edge attention estimation, and 3) relative displacement estimation.
We discuss each part in the remainder of this section.




\subsection{Keypoint Encoder}\label{sec:node embeddings}
The network first generates feature representations of the spatial layout of instances in the single-graphs, learning concise representations of their local geometry and appearance. To achieve this, our \gls{gnn} encodes information from local neighbourhoods, generating an embedding representation for each node. We limit the message passing to solely the sets of edges $E_k$ and $E_l$, preventing information exchange between the edges that form the cross-graph structure.

We initialise the feature vector of each node, denoted as $n_i \in N_k$ and $n_j \in N_l$, with its corresponding 3D spatial coordinates. Any additional semantic and feature characterisation information is solely used to inform the graph structure and gets disregarded at this step, effectively converting the graphs into homogeneous representations.
The feature vectors are then processed through the edges in the direction defined by the graph topology, as described in \cref{sec:graph_construction}.

We employ \glspl{gcn} \cite{kipf2016semi} to aggregate the feature vectors of the nodes in each graph, followed by \glspl{mlp} to process them further and learn local geometric representations. To restrain message passing to the local neighbourhood, we limit the number of \gls{gcn} layers in our implementation to two, aggregating on a spatial level.

This process generates a new cross-graph structure, denoted as $\Gamma^e_{k, l} = \langle (N^e_k \cup N^e_l), (E_k \cup E_l \cup E_{k,l}) \rangle$. While maintaining the same topological structure as $\Gamma_{k, l}$, the nodes now correspond to learned embedded keypoint representations, calculated from the \gls{gcn} and \gls{mlp} layers.


\subsection{Cross-Attention Module}\label{sec:cross_graph_attention}
\label{sec:xgraphAttn}

Given the new structure $\Gamma_{k, l}^e$, we now discard the edges connecting the nodes inside the single-graph structures and only keep the edges forming the cross-graph, denoted as $E_{k, l}$. We topologically limit the node feature vector aggregation to nodes corresponding to registration candidates. This process forms discrete subgraphs with each node in $G_l$ being connected to multiple nodes in $G_k$, as described in \cref{sec:graph_pair}, corresponding to its potential matches.

Similar to MDGAT \cite{shi2021keypoint}, we exploit attentional aggregation \cite{velivckovic2017graph} assigning weights to the edges between registration candidates.
We adopt two \gls{gat} \cite{brody2021attentive} layers alternated with a \gls{mlp} extracting the final attention coefficients for each edge as edge weights from the final \gls{gat} layer. We define such weights as $W_{k,l}: \set{w_{i,j} \given w_{i,j} \in \mathbb{R} ~\forall e_{i, j} \in E_{k, l}}$.



\subsection{Relative Displacement Estimation}\label{sec:displacement_estimation}
\label{sec:svd}

These final edge-attention scores $W_{k,l}$ correspond to the confidence score for each candidate match between $P_k$ and $P_l$, and are used as weights in \gls{svd} to estimate the displacement between $P_k$ and $P_l$.

Defining the centroids of $P_k$ and $P_l$ as: 
\begin{equation}
    \bar{\mathbf{p}}_k= {1 \over | P_k |} \sum_{\mathbf{p}_i \in P_k}\mathbf{p}_i\qquad \text{and}\qquad \bar{\mathbf{p}}_l= {1 \over |P_l|} \sum_{\mathbf{p}_i \in P_l} \mathbf{p}_i
\end{equation}
we obtain the cross-covariance matrix $H$ for \begin{equation}
    \mathbf{H} =  \sum_{e_{i,j} \in E_{k, l}} w_{i,j} (\mathbf{p}_i - \bar{\mathbf{p}_k})(\mathbf{p}_j-\bar{\mathbf{p}_l})^T
\end{equation}
where $p_i \in P_k$ and $p_j \in P_l$ and $e_{i, j}$ is the edge connecting them.
We then retrieve the transformation between $P_k$ and $P_l$ as:
\begin{equation}
    \mathbf{\hat{R}}_{k,l} = V U^\top\qquad \text{and}\qquad
    \mathbf{\hat{t}}_{k,l} = \bar{\mathbf{p}}_l - \mathbf{\hat{R}}_{k,l} \bar{\mathbf{p}}_k
\end{equation}

Our ablation study demonstrated that our model achieves higher accuracy and training stability when using only the edge from each $\mathbf{p}_j \in P_l$ with the maximum weight value.
We denote this subset of edges as $\bar{W}_{k, l}$.

\subsection{Losses}\label{sec:losses}
\label{sec:losses}

Our network is trained using a 2-component loss: $L_{a}$ applied to the attention weights and $L_p$ to the relative pose estimation output from \gls{svd}. The total loss is the sum of those two losses defined as $L_{total} = L_{a} + L_p$.

\subsubsection{Attention loss}
Given the ground-truth poses from the KITTI odometry dataset \cite{6248074}, we transform the pointcloud $P_k$ and align it with $P_l$.
We then extract ground-truth registration matches finding the closest point $\mathbf{p}_j \in P_l$ to each point $\mathbf{p}_i \in P_k$ within a 2~\si{\meter} region.
We empirically chose the optimal threshold value that minimises the pose error after applying \gls{svd} on the computed ground truth matches.

We then use these ground-truth correspondences in a classification setting employing a binary cross-entropy loss. Given $W_{gt} \in \{0,1\}$, the ground-truth weights corresponding to the ground-truth registration matches, and $\bar{W}_{k, l} \in [0,1]$, the predicted attention weights from our model, we calculate the attention loss as:
\begin{equation}
\begin{split}
    &L_a = -w_n(W_{gt}\log(\bar{W}_{k, l}) + (1 - W_{gt})\log(1 - \bar{W}_{k, l})) \\
    \end{split}
\end{equation}
where $w_n$ corresponds to the positive weight calculated as the ratio between the number of negative and positive labels in the ground truth weights, applied to mitigate class imbalance.

\subsubsection{Pose Estimation loss}
Following the relative displacement estimation module described in \cref{sec:displacement_estimation}, we compute the transformation loss $L_p$ by calculating the translation loss $L_t$ and the rotation loss $L_r$ as follows:
\begin{equation}
L_p = \alpha L_r + L_t
\end{equation}
where $\alpha$ corresponds to a scaling factor, in our case $1\mathrm{e}3$, and
\begin{equation}
L_t = ||\mathbf{t}_{gt} - \hat{\mathbf{t}}|| \qquad \mathrm{and} \qquad
L_r = \mathrm{tr}(\mathbf{I} - \mathbf{R}^\top_{gt}\hat{\mathbf{R}})
\end{equation}
Here, $\mathbf{t}_{gt}$ and $\mathbf{R}_{gt}$ are the ground truth translation and rotation from KITTI odometry dataset \cite{6248074}, and $\hat{\mathbf{t}}$ and $\hat{\mathbf{R}}$ are the predicted translation and rotation from SEM-GAT.

\section{Experimental Setup} \label{sec:experimental_setup}
\subsection{Implementation Details}\label{sec:implementation_details}
We evaluate our method on the SemanticKITTI dataset \cite{Behley2019}, using the ground-truth poses and labels provided considering all pairs of consecutive scans. Sequences 00, 02, and 03 are used with an 80--20 \% split as training and validation sets respectively. We test our method on the remaining sequences.

After iterative experimentation and refinement, we concluded on the optimal setup for our approach. We first map the dynamic classes to their associated static ones and discard \texttt{road} and \texttt{parking} due to their poor quality and inconsistent labelling, achieving improved performance. We extract the instances from each semantic class applying the cluster sizes -- e.g. $50$ for traffic sign as opposed to $200$ for vegetation -- as well as different maximum distances between neighbouring points within a cluster -- e.g. $0.5$ for car as opposed to $2$ for building or terrain, as proposed by \cite{Kong2020sgpr}.
We build the graph with a \SI{0.8}{\metre} distance threshold for nearest neighbour connections for feature nodes in the same instance on the single-graphs. We experimentally concluded the optimal distance thresholds for point-to-point registration candidates and corresponding instances in the graph pairs to be 2~\si{\metre} in inference and empirically observed an improvement in performances with a larger (3~\si{\metre}) threshold during training.

In the SEM-GAT network, for the first message passing step of the node features aggregation module at \cref{sec:node embeddings} we use a \textit{$GCN_{f0}$} and an \textit{$MLP_{f0}$} of sizes $(3,32)$ and $(32,64,128)$ respectively. On the second message passing step for close neighbourhood node feature vector aggregation, we use a \textit{$GCN_{f1}$} and an \textit{$MLP_{f1}$} of sizes $(128,256)$ and $(256,256,256)$. To each GCN we apply a 10\% dropout. For the attention aggregation module on the cross-graph described in \cref{sec:xgraphAttn} we use 3 attention heads for the first \textit{$GAT_{f0}$} of size $(256, 128)$ followed by an \textit{$MLP_{f3}$} of size $(128 * 3, 64, 32)$. At the final step, in which we estimate the attention scores, we use a \textit{$GAT_{f1}$} of size $(32, 8)$ with one attention head. The scores are then used in the weighted \gls{svd} as described in \cref{sec:svd}.
We train the network with the Adam optimiser with a learning rate of $1\mathrm{e}-3$ for 80 epochs with early termination and batch size of 4 on an NVIDIA GeForce RTX 2080Ti GPU.

\subsection{Evaluation Metrics}\label{sec:evaluation}
To evaluate our method, we use the error metrics of \gls{rre} [\si{\degree}] and \gls{rte} [\SI{}{\metre}], defined as:
\begin{equation}
\text{RRE} = \acos(\frac{1}{2} (\mathrm{tr}(\mathbf{R}^\top_{gt}\hat{\mathbf{R}}) - 1))
\end{equation}

\begin{equation}
\text{RTE} = \norm{\mathbf{t}_{gt} - \hat{\mathbf{t}}}_2
\end{equation}
We further calculate the registration recall $\text{RR}$~[\SI{}{\percent}] measuring the percentage of registrations with rotation and translation errors within predefined thresholds, which quantifies the number of ``successful registrations''.
We follow \cite{cao2021pcam} and set $\text{RTE} < \SI{0.6}{\metre}$ and $\text{RRE} < \SI{5}{\degree}$; we report both RRE and RTE averaged over successful registrations. Results are shown in \cref{tab:rr}.

\subsection{Baselines}

We compare SEM-GAT against state-of-the-art methods such as \gls{ransac}-based dense point matching of the traditional FPFH~\cite{rusu2009fpfh} and learned FCGF~\cite{choy2019fcgf} descriptors, and the end-to-end method DGR~\cite{choy2020dgr}.
We use the pre-trained models on KITTI sequences 00-05 provided by the authors and follow the default settings, particularly downsampling input pointclouds using voxel size of 0.3 \si{\metre}.
For DGR we include the results both of the base approach with \gls{ransac} safeguard, and the most accurate version with \gls{icp} refinement as per~\cite{choy2020dgr}.
We also compare against the graph-attention approach MDGAT, pre-trained on sequences 00-07 and 09~\cite{shi2021keypoint}.
Since FCGF and DGR are trained on sequences 00-05, we only report the results on the remaining sequences 06-10.
\begin{figure*}[ht]
\centering
\includegraphics[width=\textwidth]{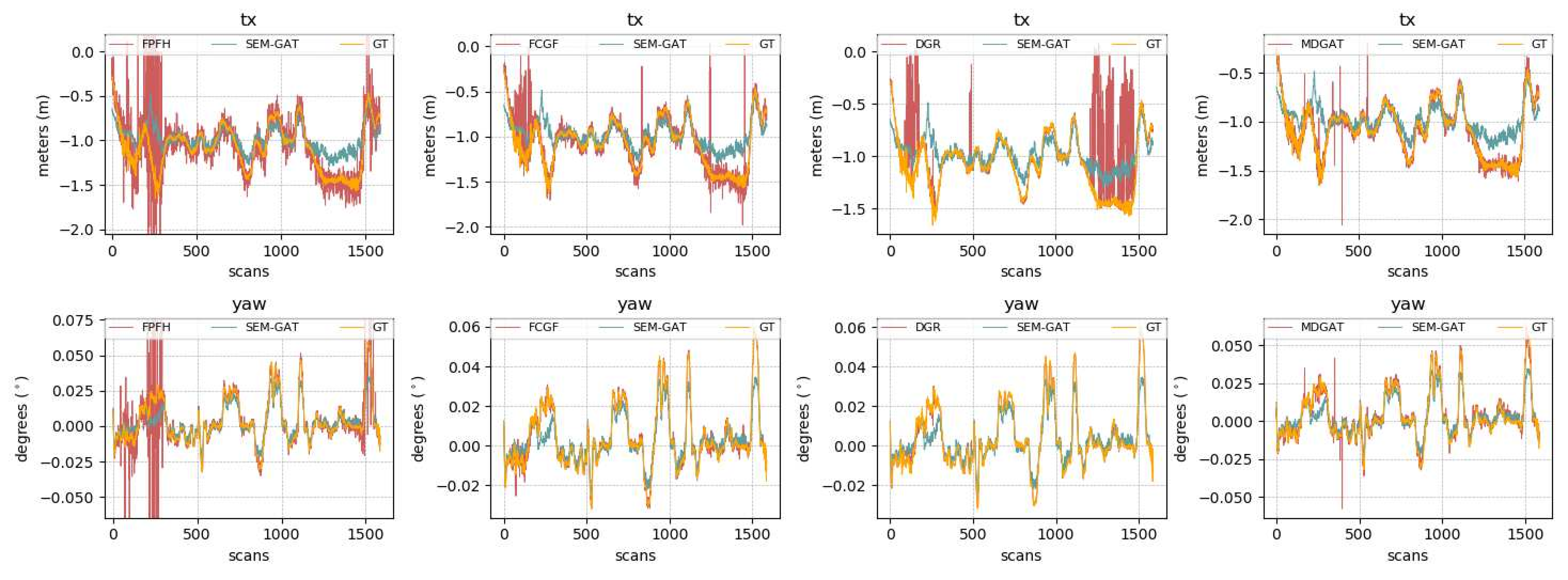}
\caption{
Transform measurements $t_x$ and $yaw$ comparison between the benchmarks, ground truth, and SEM-GAT (Ours) for sequence 09. Even though this is the most challenging sequence for our model, we still achieve more reliable performance compared to the benchmarks.
}
\label{fig:translation}
\vspace{-0.5cm}
\end{figure*}
\section{Results}

\subsection{Relative pose estimation}\label{sec:results_relative_pose}
As illustrated in \cref{tab:rr}, our registration recall performance is consistently either better or comparable to the benchmarks in every sequence, apart from sequence 09. 
For thresholds of $\text{RTE} < \SI{0.6}{\metre}$ and $\text{RRE} < \SI{5}{\degree}$, we achieve almost-perfect performance across the evaluation sequences. This is justified qualitatively in \cref{fig:translation}.
The relative pose measurements estimated from FPFH, DGR, MDGAT, and FCGF return significantly large errors in certain areas of the pointcloud, seen as spikes in \cref{fig:translation}, making them unreliable for an odometry pipeline.
This is the case for DGR even with the presence of its \gls{ransac} safeguard, which acts as a fallback whenever the confidence of DGR's base estimate is low.
Even though the average translation and rotation errors of the benchmarks are low, our method returns more stable and smooth measurements making the overall pose estimation system more robust and reliable in a downstream odometry pipeline.
Compared to the benchmarks, we achieve robust overall performance, competitive with the recent state-of-the-art.

As discussed in \cref{sec:graph_pruning}, our graphs are extremely lightweight compared to the ones from methods relying on fully-connected structures. 
Since the graph acts as an inductive bias for data aggregation, our network can achieve competitive performances at a fraction of the learnt parameters.
\Cref{tab:nn-params} shows the number of parameters in our network, where SEM-GAT is far more lightweight than comparable methods, while still achieving good accuracy results. 

\begin{table}[h]
\centering
\renewcommand{\arraystretch}{1.2}
\begin{tabular}{c|cc}
\textbf{Model} & \textbf{Parameters} $\downarrow$ & vs SEM-GAT $\downarrow$ \\
\hline
MDGAT & 3045k & $6\times$ \\
FCGF & 8753k & $18\times$ \\
DGR & 245M & $514\times$ \\
\hline
SEM-GAT (Ours) & \textbf{476k} & $\mathbf{1\times}$ \\
\end{tabular}
\caption{Network parameter count compared to benchmarks.}\label{tab:nn-params}
\vspace{-0.5cm}
\end{table}

\begin{table*}[h]
\centering
\resizebox{\textwidth}{!}{
\renewcommand{\arraystretch}{1.2}
\begin{tabular}{l|ccc|ccc|ccc|ccc|ccc}
Method & \multicolumn{3}{c|}{06} & \multicolumn{3}{c|}{07} & \multicolumn{3}{c|}{08} & \multicolumn{3}{c}{09} & \multicolumn{3}{c}{10} \\
& RR & RTE & RRE &  RR & RTE & RRE &  RR & RTE & RRE &  RR & RTE & RRE &  RR & RTE & RRE \\

\hline
FPFH & \textbf{100} & \num{0.11590509} & \num{0.19098033} & \textbf{100} & \num{0.11423229} & \num{0.28362937} & \textbf{100} & \num{0.11688976} & \num{0.27216883} & \num{99.119497} & \num{0.12299306} & \num{0.30756113} & \num{99.833333} & \num{0.11241671} & \num{0.34460725} \\
FCGF & \textbf{100} & \num{0.070099278} & \num{0.11781415} & \textbf{100} & \num{0.079428967} & \num{0.181136524} & \num{99.9017199} & \num{0.086684851} & \num{0.194271773} & \num{98.5534591} & \num{0.095335051} & \num{0.223971575} & \num{99.4166667} & \num{0.087098085} & \num{0.235996558} \\
DGR (no ICP) & \num{58.90909090909090} & \num{0.36505933004753} & \num{0.183671658037786} & \num{70} & \num{0.283255550244493} & \num{0.125644471165556} & \num{59.2628992628993} & \num{0.318311255877088} & \num{0.153136600802605} & \num{41.3207547169811} & \num{0.412188399567312} & \num{0.200164387699697} & \num{60.75} & \num{0.331413435644153} & \num{0.185415758538093} \\
DGR & \num{99.7272727} & \textbf{\num{0.027900708}} & \textbf{\num{0.059699367}} & \num{98.1818182} & \textbf{\num{0.030772097}} & \textbf{\num{0.083404333}} & \num{99.8034398} & \textbf{\num{0.033302616}} & \textbf{\num{0.082109148}} & \num{90.4402516} & \textbf{\num{0.043617187}} & \textbf{\num{0.087130238}} & \num{95.25} & \textbf{\num{0.034178421}} & \textbf{\num{0.103303595}} \\
MDGAT & \num{99.9090082} & \num{0.059226077} & \num{0.139241653} & \num{99.0900819} & \num{0.051816425} & \num{0.198455705} & \num{99.4101745} & \num{0.062000004} & \num{0.203497035} & \num{99.8740554} & \num{0.056197864} & \num{0.23103047} & \num{99.1659716} & \num{0.058182021} & \num{0.246501083} \\
\hline
Ours & \textbf{100} & \num{0.182228951} & \num{0.254088552} & \textbf{100} & \num{0.135788435} & \num{0.2302207} & \textbf{100} & \num{0.125242852} & \num{0.270728138} & \num{98.7421384} & \num{0.149581416} & \num{0.291431535} & \textbf{100} & \num{0.148216158} & \num{0.320466947} \\
\end{tabular}
}
\caption{
Per sequence
RR [\SI{}{\percent}], $\text{RTE} < \SI{0.6}{\metre}$ and $\text{RRE} < \SI{5}{\degree}$ on the KITTI odometry dataset~\cite{6248074}.
}\label{tab:rr}
\vspace{0.2cm}
\end{table*}

\begin{table*}[h]
\resizebox{\textwidth}{!}{
\renewcommand{\arraystretch}{1.2}
\begin{tabular}{l|ccccccccccc|c||cc}
Class             & 00               & 01               & 02               & 03               & 04               & 05               & 06               & 07               & 08               & 09               & 10               & Total             & corner          & surface         \\
                  & Urban            & Highway          & Urban            & Country          & Country          & Country          & Urban            & Urban            & Urban            & Urban            & Country          & & & \\
\hline
sidewalk          & \num{0.53252156} & \num{0}          & \textbf{\num{0.56067729}} & \num{0.54545636} & \textbf{\num{0.59687717}} & \textbf{\num{0.55859932}} & \num{0.56963359} & \num{0.53942218} & \textbf{\num{0.54666415}} & \textbf{\num{0.55181193}} & \num{0.44011246} & \num{0.49470691}  & \num{0.54155649} & \num{0.36266608} \\
fence             & \num{0.43526244} & \textbf{\num{0.51218543}} & \num{0.483654}   & \num{0.51149344} & \num{0.44798557} & \num{0.46522949} & \num{0.39689758} & \num{0.45656687} & \num{0.29228247} & \num{0.38376786} & \num{0.46593493} & \num{0.44102364}  & \num{0.4841983}  & \num{0.41972867} \\
pole              & \textbf{\num{0.55257704}} & \num{0.05718674} & \num{0.36392915} & \textbf{\num{0.5509522}}  & \num{0.493527}   & \num{0.29214354} & \textbf{\num{0.59823658}} & \textbf{\num{0.56441706}} & \num{0.50871846} & \num{0.36810948} & \textbf{\num{0.49035}}    & \num{0.44001339}  & \num{0.37178395} & \num{0.37540313} \\
vegetation        & \num{0.38444364} & \num{0.42439203} & \num{0.39549307} & \num{0.37832707} & \num{0.42540497} & \num{0.37064093} & \num{0.40202463} & \num{0.38166452} & \num{0.38179694} & \num{0.39087427} & \num{0.37619137} & \num{0.39193213}  & \num{0.43638689} & \num{0.34814316} \\
terrain           & \num{0.2922082}  & \num{0.38954782} & \num{0.30858521} & \num{0.38152513} & \num{0.41902206} & \num{0.49542731} & \num{0.44681112} & \num{0.20245937} & \num{0.43138961} & \num{0.43938085} & \num{0.29180537} & \num{0.37256019}  & \num{0.42969453} & \num{0.25399524} \\
car               & \num{0.39272067} & \num{0.28870442} & \num{0.33919202} & \num{0.2981226}  & \num{0.43616554} & \num{0.40342127} & \num{0.42890466} & \num{0.39027927} & \num{0.40291613} & \num{0.33886028} & \num{0.34747198} & \num{0.36970535}  & \num{0.38072539} & \num{0.36261918} \\
trunk             & \num{0.34045464} & \num{0.02510803} & \num{0.45197305} & \num{0.34145598} & \num{0.32916694} & \num{0.3138616}  & \num{0.51843915} & \num{0.29333243} & \num{0.4194015}  & \num{0.43022768} & \num{0.25244762} & \num{0.33780624}  & \num{0.32919182} & \num{0.30229021} \\
building          & \num{0.4000298}  & \num{0.03068569} & \num{0.33354435} & \num{0.27931574} & \num{0.195332}   & \num{0.40322825} & \num{0.41881818} & \num{0.39557093} & \num{0.39718872} & \num{0.3462839}  & \num{0.36722016} & \num{0.32429252}  & \num{0.41896654} & \num{0.28325162} \\
traffic sign      & \num{0.08437592} & \num{0.09861421} & \num{0.05287499} & \num{0.1532084}  & \num{0.18118766} & \num{0.08073686} & \num{0.3454745}  & \num{0.09883529} & \num{0.12210132} & \num{0.12168991} & \num{0.08762822} & \num{0.12970248}  & \num{0.10014248} & \num{0.11435768} \\
other-vehicle     & \num{0.07292595} & \num{0.01645499} & \num{0.02718834} & \num{0.0607798}  & \num{0.05054882} & \num{0.09852098} & \num{0.39754044} & \num{0.09745809} & \num{0.15566219} & \num{0.07497168} & \num{0.03912742} & \num{0.09919806}  & \num{0.10566432} & \num{0.09060917} \\
bicycle           & \num{0.40340226} & \num{0}          & \num{0.00594349} & \num{0.04866676} & \num{0}          & \num{0.04074344} & \num{0.14454348} & \num{0.10962389} & \num{0.17821615} & \num{0.01063271} & \num{0.01038066} & \num{0.08655935}  & \num{0.05481024} & \num{0.03578428} \\
person            & \num{0.07188454} & \num{0}          & \num{0.02334716} & \num{0.03004875} & \num{0.04410702} & \num{0.08397642} & \num{0.08269778} & \num{0.18761185} & \num{0.16540963} & \num{0.10146784} & \num{0.10020192} & \num{0.08097754}  & \num{0.07519542} & \num{0.06109099} \\
other-ground      & \num{0}          & \num{0.17722161} & \num{0.01998006} & \num{0}          & \num{0.10045647} & \num{0.03688137} & \num{0.26510175} & \num{0}          & \num{0.01906919} & \num{0.03712971} & \num{0.12591108} & \num{0.07106829}  & \num{0.0788743}  & \num{0.04589642} \\
motorcycle        & \num{0.07946423} & \num{0}          & \num{0.04715691} & \num{0}          & \num{0}          & \num{0.00931199} & \num{0.08428278} & \num{0.0730609}  & \num{0.07113503} & \num{0.04683634} & \num{0.04054257} & \num{0.04107189}  & \num{0.03857563} & \num{0.03230282} \\
truck             & \num{0.0157399}  & \num{0}          & \num{0}          & \num{0}          & \num{0}          & \num{0.05257322} & \num{0.15715448} & \num{0.09296475} & \num{0.01628239} & \num{0.04443665} & \num{0.02832584} & \num{0.03704338}  & \num{0.04386739} & \num{0.03257054} \\
\hline
corners           & \num{0.2655345}  & \num{0.2176132}  & \num{0.26271385} & \num{0.31470257} & \num{0.33089762} & \num{0.25839367} & \num{0.37529273} & \num{0.28613698} & \num{0.28516739} & \num{0.26105975} & \num{0.24297843} & \num{0.28186279}  & /                & /                \\
surfaces          & \num{0.2109026}  & \num{0.17725493} & \num{0.21352665} & \num{0.26176217} & \num{0.26331079} & \num{0.20946892} & \num{0.29408772} & \num{0.22339295} & \num{0.23137805} & \num{0.21124974} & \num{0.19193075} & \num{0.22620593}  & /                & /                \\     
\end{tabular}
}
\caption{Average learned edge attention categorised by class, presented per SemanticKITTI class and sorted over all sequences (\textit{Total}).}\label{tab:edge_weights}
\end{table*}

\subsection{Explainability Analysis}
\label{sec:results_introspect}
\begin{figure}[h]
\centering
\includegraphics[width=0.6\columnwidth]{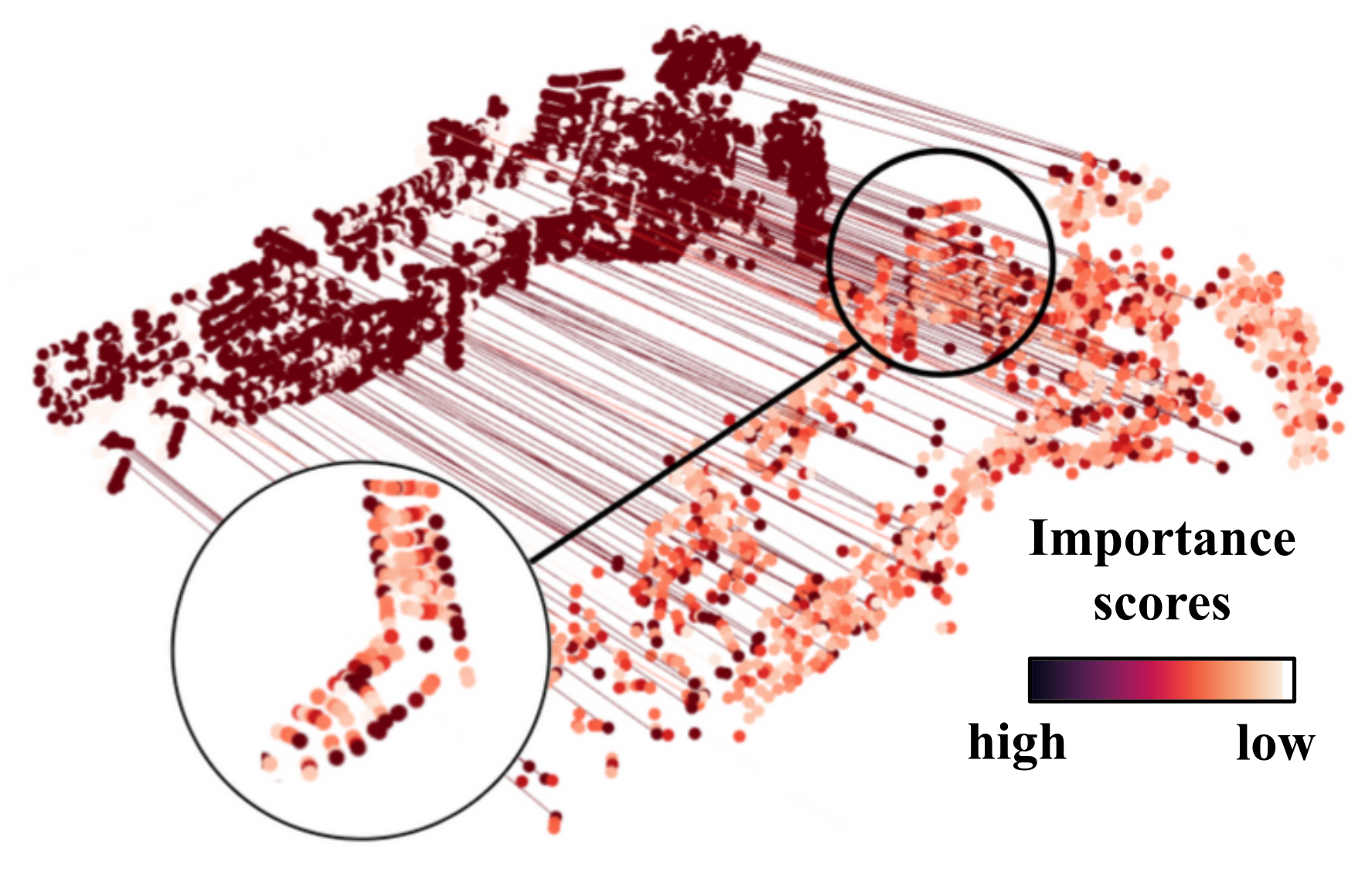}
\caption{\label{fig:introspectionEdgesResult}
Heatmap representation of edge attention weight distribution on the nodes of the corresponding pointcloud. 
}
\end{figure}

As our network estimates pointcloud displacement directly from graph edges by assigning them confidence scores, we can examine node importance with respect to each match's influence on the behaviour of our model after inspecting such score. By calculating the accumulated average attention weights from each semantic and geometric class, we evaluate their individual and combined contributions to the performance of our pose estimation pipeline.
This is illustrated in \cref{tab:edge_weights} and visualised through an example in \cref{fig:introspectionEdgesResult}. 

We conclude that the network consistently prioritises \texttt{pole} and \texttt{sidewalk} semantic classes regardless of whether the pointcloud corresponds to an urban, country, or highway environment. The network further prioritises \texttt{surface} and \texttt{corner} points according to the geometric properties of each semantic class. Based on the output reported in \cref{tab:edge_weights}, in every semantic class, apart from \texttt{pole}, the network prioritises corner points instead of surface points, especially in classes rich in flat surfaces like \texttt{sidewalk}. If we correlate these estimations with the reduced performance of our network on long straight roads, like the ones in Sequence 09 visualised in \cref{fig:translation}, we observe that the registration performance drops when the morphology of the environment is relatively repetitive.
This analysis gives a thorough understanding of the behaviour of our model in correlation with the environment, allowing us to anticipate its behaviour across diverse morphological domains.

\section{Ablation}\label{sec:ablation}
We analyse the effects of the two losses and our architectural choices through experimental evaluation.
We use the validation split of the training set to evaluate the accuracy of our network in terms of \gls{mse} loss while training on the corresponding training split. We tried different network configuration settings including various architectures, convolutions, network sizes, aggregation and attention modules, and eventually losses.
Representative results of the best performance obtained from various configurations in our ablation study can be seen in ~\cref{tab:ablationtable}.
We observe the contribution of both the BCE and MSE losses, as well as the relevance of the depth of the attention module.
This study was complemented by hyperparameter tuning on the nearest neighbours and corresponding distances to conclude on the most robust graph configuration.
\begin{table}[h]
\centering
\begin{tabular}{|p{0.1\linewidth}|p{0.1\linewidth}|p{0.13\linewidth}|p{0.1\linewidth}|p{0.12\linewidth}|}
\hline
BCE  Loss & MSE  Loss & Attention  Heads &  MLP layers &  MSE~$\mathbf{\downarrow}$ \\ \hline
- & \checkmark & 3 & 3 & 0.5 \\ 
- & \checkmark & 1 & 2 & 0.4 \\ 
- & \checkmark & 1 & 1 & 0.14 \\ 
\checkmark & - & 2 & 2 & 0.29 \\ 
\checkmark & \checkmark & 2 & 1 & 0.35 \\ 
\checkmark & \checkmark & 3 & 0 & 0.26 \\ 
\checkmark & \checkmark & 3 & 3 & \textbf{0.084} \\ 
\hline
\end{tabular}
\caption{Ablation study results for architectural search.}
\label{tab:ablationtable}
\vspace{-0.2cm}
\end{table}

\section{Conclusions}\label{sec:conclusion}
In this work, we presented a novel \gls{gnn}-based approach for lidar pose estimation. We introduced a lightweight graph representation structure relying on scene morphology and semantics to guide the learning and inference process. For keypoint extraction, we aggregate geometric information within local structures defined by graph topology to generate strong registration candidates. Our model relies on graph attention to identify the final matches and align the pointclouds estimating the relative displacement between them. Our model achieves higher and more reliable performance, compared to the benchmarks, successfully estimating the pose between scans. Analysing the output from our attention module, we explain the performance of the model in correlation with the semantics and the morphology of the environment. This gives us a thorough understanding of the behaviour of our model in various domains. The structure of our entire pipeline is such that allows high-level introspection which is a crucial component of eXplainable AI (XAI).

\bibliographystyle{ieeetr}
\bibliography{biblio}

\begin{thebibliography}{10}

\bibitem{He2016M2DP}
L.~He, X.~Wang, and H.~Zhang, ``{M2dp: A novel 3D point cloud descriptor and
  its application in loop closure detection},'' {\em IEEE International
  Conference on Intelligent Robots and Systems}, vol.~2016-Novem, pp.~231--237,
  2016.

\bibitem{rusu2009fpfh}
R.~B. Rusu, N.~Blodow, and M.~Beetz, ``Fast point feature histograms (fpfh) for
  3d registration,'' in {\em 2009 IEEE International Conference on Robotics and
  Automation}, pp.~3212--3217, 2009.

\bibitem{sun2020dagc}
Q.~Sun, H.~Liu, J.~He, Z.~Fan, and X.~Du, ``{DAGC}: Employing dual attention
  and graph convolution for point cloud based place recognition,'' in {\em
  Proceedings of the 2020 International Conference on Multimedia Retrieval},
  pp.~224--232, 2020.

\bibitem{Zhang2019pcan}
W.~Zhang and C.~Xiao, ``{PCAN: 3D attention map learning using contextual
  information for point cloud based retrieval},'' {\em Proceedings of the IEEE
  Computer Society Conference on Computer Vision and Pattern Recognition},
  vol.~2019-June, pp.~12428--12437, 2019.

\bibitem{Dube2017SegMatch}
R.~Dube, D.~Dugas, E.~Stumm, J.~Nieto, R.~Siegwart, and C.~Cadena, ``{SegMatch:
  Segment based place recognition in 3D point clouds},'' {\em Proceedings -
  IEEE International Conference on Robotics and Automation}, pp.~5266--5272,
  2017.

\bibitem{Li2021b}
L.~Li, X.~Kong, X.~Zhao, W.~Li, F.~Wen, H.~Zhang, and Y.~Liu, ``Sa-loam:
  Semantic-aided lidar slam with loop closure,'' in {\em 2021 IEEE
  International Conference on Robotics and Automation (ICRA)}, pp.~7627--7634,
  2021.

\bibitem{Wang2019DCP}
Y.~Wang and J.~Solomon, ``{Deep closest point: Learning representations for
  point cloud registration},'' {\em Proceedings of the IEEE International
  Conference on Computer Vision}, vol.~2019-Octob, pp.~3522--3531, 2019.

\bibitem{fischer2021stickypillars}
K.~Fischer, M.~Simon, F.~Olsner, S.~Milz, H.-M. Gross, and P.~Mader,
  ``Stickypillars: Robust and efficient feature matching on point clouds using
  graph neural networks,'' in {\em Proceedings of the IEEE/CVF Conference on
  Computer Vision and Pattern Recognition}, pp.~313--323, 2021.

\bibitem{Li2021SSC}
Y.~Li, P.~Su, M.~Cao, H.~Chen, X.~Jiang, and Y.~Liu, ``{Semantic scan context:
  Global semantic descriptor for LiDAR-based place recognition},'' {\em 2021
  IEEE International Conference on Real-Time Computing and Robotics, RCAR
  2021}, pp.~251--256, 2021.

\bibitem{6248074}
A.~Geiger, P.~Lenz, and R.~Urtasun, ``Are we ready for autonomous driving? the
  kitti vision benchmark suite,'' in {\em 2012 IEEE Conference on Computer
  Vision and Pattern Recognition}, pp.~3354--3361, 2012.

\bibitem{besl1992method}
P.~J. Besl and N.~D. McKay, ``Method for registration of 3-d shapes,'' in {\em
  Sensor fusion IV: control paradigms and data structures}, vol.~1611,
  pp.~586--606, Spie, 1992.

\bibitem{Rusinkiewicz2001}
S.~Rusinkiewicz and M.~Levoy, ``{Efficient variants of the ICP algorithm},''
  {\em Proceedings of International Conference on 3-D Digital Imaging and
  Modeling, 3DIM}, pp.~145--152, 2001.

\bibitem{Segal2009}
A.~V. Segal, D.~Haehnel, and S.~Thrun, ``{Generalized-ICP (probailistic ICP
  tutorial)},'' {\em Robotics: science and systems}, vol.~2, p.~435, 2009.

\bibitem{Zhang2022}
J.~Zhang, Y.~Yao, and B.~Deng, ``{Fast and Robust Iterative Closest Point},''
  {\em IEEE Transactions on Pattern Analysis and Machine Intelligence},
  vol.~44, no.~7, pp.~3450--3466, 2022.

\bibitem{Mendes2016}
E.~Mendes, P.~Koch, and S.~Lacroix, ``Icp-based pose-graph slam,'' in {\em 2016
  IEEE International Symposium on Safety, Security, and Rescue Robotics
  (SSRR)}, pp.~195--200, 2016.

\bibitem{Borrmann2008}
D.~Borrmann, J.~Elseberg, K.~Lingemann, A.~N{\"{u}}chter, and J.~Hertzberg,
  ``{Globally consistent 3D mapping with scan matching},'' {\em Robotics and
  Autonomous Systems}, vol.~56, no.~2, pp.~130--142, 2008.

\bibitem{li2019usip}
J.~Li and G.~H. Lee, ``Usip: Unsupervised stable interest point detection from
  3d point clouds,'' in {\em 2019 IEEE/CVF International Conference on Computer
  Vision (ICCV)}, pp.~361--370, 2019.

\bibitem{choy2019fcgf}
C.~Choy, J.~Park, and V.~Koltun, ``Fully convolutional geometric features,'' in
  {\em 2019 IEEE/CVF International Conference on Computer Vision (ICCV)},
  pp.~8957--8965, 2019.

\bibitem{rusu20113d}
R.~B. Rusu and S.~Cousins, ``3d is here: Point cloud library (pcl),'' in {\em
  2011 IEEE international conference on robotics and automation}, pp.~1--4,
  IEEE, 2011.

\bibitem{yang2021teaser}
H.~Yang, J.~Shi, and L.~Carlone, ``Teaser: Fast and certifiable point cloud
  registration,'' {\em IEEE Transactions on Robotics}, vol.~37, no.~2,
  pp.~314--333, 2021.

\bibitem{Fischler1981}
M.~A. Fischler and R.~C. Bolles, ``{RANSAC: Random Sample Paradigm for Model
  Consensus: A Apphcatlons to Image Fitting with Analysis and Automated
  Cartography},'' {\em Graphics and Image Processing}, vol.~24, no.~6,
  pp.~381--395, 1981.

\bibitem{Wang2019c}
Y.~Wang and J.~Solomon, ``{Deep closest point: Learning representations for
  point cloud registration},'' {\em Proceedings of the IEEE International
  Conference on Computer Vision}, vol.~2019-Octob, pp.~3522--3531, 2019.

\bibitem{Li2020b}
J.~Li, C.~Zhang, Z.~Xu, H.~Zhou, and C.~Zhang, ``{Iterative Distance-Aware
  Similarity Matrix Convolution with Mutual-Supervised Point Elimination for
  Efficient Point Cloud Registration},'' {\em Lecture Notes in Computer Science
  (including subseries Lecture Notes in Artificial Intelligence and Lecture
  Notes in Bioinformatics)}, vol.~12369 LNCS, pp.~378--394, 2020.

\bibitem{Lu2019}
W.~Lu, G.~Wan, Y.~Zhou, X.~Fu, P.~Yuan, and S.~Song, ``{DeepICP: An end-to-end
  deep neural network for point cloud registration},'' {\em Proceedings of the
  IEEE International Conference on Computer Vision}, vol.~2019-October,
  pp.~12--21, 2019.

\bibitem{choy2020dgr}
C.~Choy, W.~Dong, and V.~Koltun, ``Deep global registration,'' in {\em 2020
  IEEE/CVF Conference on Computer Vision and Pattern Recognition (CVPR)},
  pp.~2511--2520, 2020.

\bibitem{cao2021pcam}
A.-Q. Cao, G.~Puy, A.~Boulch, and R.~Marlet, ``Pcam: Product of cross-attention
  matrices for rigid registration of point clouds,'' in {\em 2021 IEEE/CVF
  International Conference on Computer Vision (ICCV)}, pp.~13209--13218, 2021.

\bibitem{qin2022geo}
Z.~Qin, H.~Yu, C.~Wang, Y.~Guo, Y.~Peng, and K.~Xu, ``Geometric transformer for
  fast and robust point cloud registration,'' in {\em 2022 IEEE/CVF Conference
  on Computer Vision and Pattern Recognition (CVPR)}, pp.~11133--11142, 2022.

\bibitem{Sarlin2020SuperGlue}
P.~E. Sarlin, D.~Detone, T.~Malisiewicz, and A.~Rabinovich, ``{SuperGlue:
  Learning Feature Matching with Graph Neural Networks},'' {\em Proceedings of
  the IEEE Computer Society Conference on Computer Vision and Pattern
  Recognition}, pp.~4937--4946, 2020.

\bibitem{shi2021keypoint}
C.~Shi, X.~Chen, K.~Huang, J.~Xiao, H.~Lu, and C.~Stachniss, ``Keypoint
  matching for point cloud registration using multiplex dynamic graph attention
  networks,'' {\em IEEE Robotics and Automation Letters}, vol.~6, no.~4,
  pp.~8221--8228, 2021.

\bibitem{Kong2020sgpr}
X.~Kong, X.~Yang, G.~Zhai, X.~Zhao, X.~Zeng, M.~Wang, Y.~Liu, W.~Li, and
  F.~Wen, ``{Semantic graph based place recognition for 3D point clouds},''
  {\em IEEE International Conference on Intelligent Robots and Systems},
  no.~September, pp.~8216--8223, 2020.

\bibitem{pramatarov2022boxgraph}
G.~Pramatarov, D.~De~Martini, M.~Gadd, and P.~Newman, ``Boxgraph: Semantic
  place recognition and pose estimation from 3d lidar,'' in {\em 2022 IEEE/RSJ
  International Conference on Intelligent Robots and Systems (IROS)},
  pp.~7004--7011, IEEE, 2022.

\bibitem{zhang2014loam}
J.~Zhang, S.~Singh, and V.~Sze, ``Loam: Lidar odometry and mapping in
  real-time,'' {\em Robotics: Science and Systems (RSS)}, vol.~2, no.~9,
  pp.~1--10, 2014.

\bibitem{joshi2020transformers}
C.~Joshi, ``Transformers are graph neural networks,'' {\em
  https://thegradient.pub/transformers-are-gaph-neural-networks/}, 2020.

\bibitem{murnanegraph}
D.~Murnane, ``Graph structure from point clouds: Geometric attention is all you
  need,'' {\em NeurIPS Workshop on Machine Learning and the Physical Sciences},
  2022.

\bibitem{Behley2019}
J.~Behley, M.~Garbade, A.~Milioto, J.~Quenzel, S.~Behnke, C.~Stachniss, and
  J.~Gall, ``{SemanticKITTI: A dataset for semantic scene understanding of
  LiDAR sequences},'' {\em Proceedings of the IEEE International Conference on
  Computer Vision}, vol.~2019-October, no.~iii, pp.~9296--9306, 2019.

\bibitem{kipf2016semi}
T.~N. Kipf and M.~Welling, ``Semi-supervised classification with graph
  convolutional networks,'' in {\em International Conference on Learning
  Representations}, 2017.

\bibitem{velivckovic2017graph}
P.~Veličković, G.~Cucurull, A.~Casanova, A.~Romero, P.~Liò, and Y.~Bengio,
  ``Graph attention networks,'' in {\em International Conference on Learning
  Representations}, 2018.

\bibitem{brody2021attentive}
S.~Brody, U.~Alon, and E.~Yahav, ``How attentive are graph attention
  networks?,'' in {\em International Conference on Learning Representations},
  2022.

\end{thebibliography}

\end{document}